\def\year{2022}\relax
\documentclass[letterpaper]{article} 
\usepackage{aaai22}  
\usepackage{times}  
\usepackage{helvet}  
\usepackage{courier}  
\usepackage[hyphens]{url}  
\usepackage{graphicx} 
\def\UrlFont{\rm}  
\usepackage{natbib}  
\usepackage{caption} 
\DeclareCaptionStyle{ruled}{labelfont=normalfont,labelsep=colon,strut=off} 
\usepackage{algorithm}
\usepackage{algorithmic}

\usepackage{booktabs}
\usepackage{tablefootnote}

\usepackage{xspace}

\newcommand{\ie}{\textit{i.e.,}\xspace}

\usepackage{newfloat}
\usepackage{listings}
\floatstyle{ruled}
\newfloat{listing}{tb}{lst}{}
\floatname{listing}{Listing}
\title{An Empirical Investigation of Multi-bridge Multilingual NMT models}
\author {
Anoop Kunchukuttan
}
\affiliations{
Microsoft India, Hyderabad \\
\texttt{ankunchu@microsoft.com}
}

\begin{document}
\maketitle
\begin{abstract}
In this paper, we present an extensive investigation of multi-bridge, many-to-many multilingual NMT models (MB-M2M) \ie  models trained on non-English language pairs in addition to English-centric language pairs. In addition to validating previous work which shows that MB-MNMT models can overcome zeroshot translation problems, our analysis reveals the following results about multibridge models: (1) it is possible to extract a reasonable amount of parallel corpora between non-English languages for low-resource languages (2) with limited non-English centric data, MB-M2M models are competitive with or outperform pivot models, (3) MB-M2M models can outperform English-Any models and perform at par with Any-English models, so a single multilingual NMT system can serve all translation directions.
\end{abstract}

\section{Introduction}

Neural Machine Translation has led to significant advances in MT quality in recent times \citep{bahdanau15,DBLP:journals/corr/WuSCLNMKCGMKSJL16,sennrich-etal-2016-neural,sennrich-etal-2016-improving,vaswani2017attention}. MT research has seen significant efforts in translation between English and other languages, driven in significant measure by availability of English-centric parallel corpora. Particularly, multilingual NMT models using English-centric parallel corpora have shown significant improvements for translation between English and low-resources languages \citep{firat16,johnson17}. Translation between non-English languages has received lesser attention, with the default approach being pivot translation \citep{lakew-etal-2017-zeroshot}. Pivot translation is a strong baseline, but needs multiple decoding steps resulting in increased latency and cascading errors. 

Zeroshot translation using English-centric many-to-many multilingual models (EC-M2M) \citep{johnson17} is promising, but is plagued by problems of spurious correlation between input and output language \citep{gu-etal-2019-improved,arivazhagan2018missing}. Hence, vanilla zeroshot translation quality significantly lags behind pivot translation. Various methods have been proposed to address these limitations by aligning encoder representations \citep{arivazhagan2018missing} or using pseudo-parallel corpus between non-English languages during training \citep{lakew-etal-2017-zeroshot}.

Recently, there has been interest in multi-bridge many-to-many multilingual models (MB-M2M, referred to as multi-bridge models henceforth). These models are trained on direct parallel corpora between non-English languages in addition to English-centric corpora \citep{rios-etal-2020-subword,freitag-firat-2020-complete,fan2020englishcentric}. Such corpora can either be mined from monolingual corpora \citep{fan2020englishcentric} using bitext mining approaches like LASER \citep{artetxe2019massively} and LABSE \citep{feng2020languageagnostic} or extracted from English-centric parallel corpora \citep{rios-etal-2020-subword,freitag-firat-2020-complete}. These works show that multi-bridge models can overcome zeroshot translation problems and perform at par/better than pivot approaches. In addition, models using separate encoders and decoders for one or more language(s) are feasible with such parallel corpora helping build modular multilingual NMT systems with modest model capacity that can be incrementally trained \citep{escolano-etal-2021-multilingual,lyu-etal-2020-revisiting}. 

\begin{table*}[]
\centering
\setlength{\tabcolsep}{3pt}

\begin{tabular}{lrrrrrrrrrrr}
\toprule
    & en    & bn    & gu    & hi    & kn    & ml    & mr    & or  & pa    & ta    & te \\   
\cmidrule(lr){2-2}
\cmidrule(lr){3-12}
bn  & 960   & 0     & 264   & 819   & 221   & 1,396 & 264   & 58  & 274   & 500   & 218   \\
gu  & 500   & 264   & 0     & 390   & 289   & 297   & 303   & 58  & 326   & 320   & 219   \\
hi  & 2,553 & 819   & 390   & 0     & 345   & 925   & 407   & 153 & 432   & 789   & 314   \\
kn  & 382   & 221   & 289   & 345   & 0     & 319   & 297   & 26  & 268   & 277   & 232   \\
ml  & 1,018 & 1,396 & 297   & 925   & 319   & 0     & 310   & 45  & 295   & 588   & 277   \\
mr  & 479   & 264   & 303   & 407   & 297   & 310   & 0     & 71  & 288   & 300   & 243   \\
or  & 180   & 58    & 58    & 153   & 26    & 45    & 71    & 0   & 76    & 79    & 39    \\
pa  & 496   & 274   & 326   & 432   & 268   & 295   & 288   & 76  & 0     & 356   & 208   \\
ta  & 1,207 & 500   & 320   & 789   & 277   & 588   & 300   & 79  & 356   & 0     & 231   \\
te  & 352   & 218   & 219   & 314   & 232   & 277   & 243   & 39  & 208   & 231   & 0     \\
\midrule
SUM & 8,127 & 4,014 & 2,466 & 4,574 & 2,274 & 4,452 & 2,483 & 605 & 2,523 & 3,440 & 1,981 \\
\cmidrule(lr){3-12}
 &  & \multicolumn{10}{c}{28,812} \\
\bottomrule
\end{tabular}
\caption{Statistics of the WAT 2021 dataset used in the experiments: English-centric and extracted non-English centric. All figures in 1000s of sentences}
\label{tab:dataset_details}
\end{table*}

In this paper, we undertake an extensive analysis of MB-M2M models to understand design choices for building improved multilingual NMT models. We focus on models trained using non-English corpora mined from English-centric corpora. The following are the major contributions of our work: 

\begin{table*}[ht]
\centering
\begin{tabular}{lrrrrrrrrrrr}
\toprule
Model                            & bn   & gu   & hi   & kn   & ml   & mr   & or   & pa   & ta   & te   & AVG \\
\midrule
Zeroshot                       & 0.4  & 0.3  & 0.3  & 0.5  & 0.6  & 0.3  & 0.2  & 0.6  & 0.4  & 0.4  & 0.4     \\
Pivot                            & 12.7 & 18.7 & 27.0 & 13.8 & 10.7 & 15.1 & 14.4 & 24.3 & 10.2 & 12.5 & 15.9    \\
\midrule
SamplePairs                      & 13.0 & 19.9 & 29.5 & 13.9 & 10.4 & 15.5 & 15.1 & 26.7 & 10.7 & 12.6 & 16.7    \\
SampleFraction                   & 12.8 & 20.3 & 30.5 & 13.9 & 10.4 & 15.8 & 15.7 & 27.1 & 10.8 & 12.8 & 17.0    \\
TrainAll                         & 12.5 & 20.2 & 30.2 & 14.1 & 10.5 & 15.5 & 15.6 & 26.9 & 10.7 & 12.7 & 16.9    \\
\midrule
\multicolumn{12}{l}{\textit{M2M model trained exclusively on Indic pairs}}              \\
SamplePairs & 11.1 & 16.7 & 18.5 & 10.5 & 8.0 & 12.4 & 12.5 & 19.1 & 8.8  & 10.0 & 12.8 \\
TrainAll                         & 12.2 & 19.6 & 29.1 & 13.3 & 9.9 & 15.2 & 14.9 & 25.9 & 10.1 & 12.3 & 16.3 \\
\bottomrule
\end{tabular}
\caption{Results for translation between non-English languages. Each cell shows the average BLEU score for translating into corresponding language. The last column shows the micro-averaged BLEU score for the model}
\label{tab:x2y_results}
\end{table*}

\begin{itemize}
\item An investigation of various sampling strategies reveals that only a limited amount of non-English parallel training data is required for improving upon pivot translation with increased non-English parallel data yielding limited gains.

\item We observe that under the proposed data sampling strategy, the MB-M2M models can outperform English-centric many-to-one and one-to-many models in any-English  and English-any directions respectively. Particularly, we observe significant gains for the challenging problem of multilingual translation from English into other languages \citep{10.1145/3406095}. Finetuning the multi-bridge models for particular directions does not result in a significant improvement. 

\item While MB-M2M models show improved translation between non-English  languages, we observe that there is a significant quality gap compared to translation in English-centric directions - pointing to the need for further research to address the gap.

\item Our experiments involve English and 10 Indian languages, unlike previous work which focussed on high-resource European languages. We show that it is possible to extract a significant amount of non-English  parallel corpora from the English-centric corpora even for these low-resource languages - indicating the feasibility of this extraction approach for low-resource languages as well. 

\end{itemize}

The rest of the paper is organized as follows. Section \ref{sec:exp} describes our experimental settings. Section \ref{sec:research_q} describes the research questions we explore. \ref{sec:results} discusses the results of our experiments. \ref{sec:conclusion} concludes the paper.

\section{Experiment Design}
\label{sec:exp}

\begin{table*}[]
\centering
\setlength{\tabcolsep}{3pt}
\begin{tabular}{lrrrrrrrrrrr}
\toprule
Model         & bn-mr       & bn-or       & bn-pa       & gu-kn       & gu-te       & kn-pa       & ml-hi       & ml-te       & mr-te       & ta-ml       & AVG \\
\midrule
bilingual              & 4.5  & 4.6  & 9.0  & 3.2  & 4.2  & 8.8  & 10.8 & 4.3  & 3.9  & 2.4 & 5.6  \\
pivot & 13.0 & 12.4 & 20.6 & 14.8 & 13.5 & 23.3 & 24.8 & 12.0 & 11.8 & 9.8 & 15.6 \\
SampleFraction  & 13.6 & 13.4 & 22.2 & 15.2 & 13.8 & 24.5 & 26.3 & 12.3 & 11.9 & 9.2 & 16.3 \\
\bottomrule
\end{tabular}
\caption{Results for translation between some non-English languages (BLEU score).}
\label{tab:x2y_sample_pairs}
\end{table*}

\begin{table}[]
\centering
\begin{tabular}{lrrrr}
\toprule
src       & labse & chrF2 & bleu & tset\_sim \\
\midrule
bn        & 81.2  & 45.8  & 14.6 & 78.1         \\
gu        & 84.4  & 51.1  & 19.0 & 83.2         \\
hi        & 86.0  & 52.9  & 19.7 & 83.8         \\
kn        & 82.8  & 47.9  & 16.4 & 81.7         \\
ml        & 83.1  & 47.3  & 16.0 & 80.1         \\
mr        & 82.6  & 47.9  & 16.2 & 81.6         \\
or        & 82.2  & 48.3  & 17.0 & 79.6         \\
pa        & 85.1  & 52.3  & 19.5 & 82.6         \\
ta        & 82.1  & 46.2  & 15.4 & 80.3         \\
te        & 83.9  & 48.2  & 16.3 & 74.5         \\
\textbf{AVG}       & \textbf{83.3}  & \textbf{48.8}  & \textbf{17.0} & \textbf{80.6}        \\
\midrule
en        & 86.9  & 53.8  & 20.9 & 80.1         \\
\bottomrule
\end{tabular}
\caption{Comparing English-centric and non-English translation quality}
\label{tab:eval_comparison}
\end{table}

\subsection{Extracting non-English centric corpora}
The parallel corpora between non-English languages is extracted from the English-centric parallel corpus using English as a pivot language. For languages $l_1, l_2$ and $e$ ($e$ being English and $l_i$ being non-English languages), given the sentence pairs $(s_{l_1},s_e)$  and $(s_{l_2},s_e)$, we mine all sentence pairs $(s_{l_1},s_{l_2})$ which are parallel to the English sentence $s_e$.

In this work, we used the WAT 2021 MultiIndicMT shared task dataset\footnote{http://lotus.kuee.kyoto-u.ac.jp/WAT/indic-multilingual} \cite{nakazawa-etal-2021-overview} containing parallel corpora from various sources between English and 10 Indian languages. From around 8.1m English-centric parallel corpora, we extract 14.3m  sentence pairs between Indic languages. The statistics of the English-centric training set and extracted Indic parallel corpora are shown in Table \ref{tab:dataset_details}. This shows that a reasonable amount of parallel corpus can be mined between low-resources languages too from English-centric parallel corpora.

\subsection{Multi-bridge Model}
We train a single multilingual model using data from English-X, X-English and X-Y language pairs using the multilingual model proposed by \citet{johnson17}. The input sequence contains a special token indicating the target language. We also include a special token to indicate the source language \citep{tan-etal-2019-multilingual,tang2020multilingual}, allowing the model to utilize the source language tag as well as the similarity of encoder representations for transfer learning. We use a subword vocabulary size of ~32K.  We use \textit{transformer\_vaswani\_wmt\_en\_de\_big} architecture as defined in the fairseq toolkit. Specific model details are mentioned below.

\subsubsection{Preprocessing}

We convert all the Indic data to the Devanagari script. This allows better lexical sharing between languages for transfer learning, prevents fragmentation of the subword vocabulary between Indic languages and allows using a smaller subword vocabulary. When the target language is Indic, the output in Devanagari script is converted back to the corresponding Indic script. Other standard pre-processing done on the data are Unicode normalization and tokenization. All Indic language text processing uses the Indic NLP library\footnote{https://github.com/anoopkunchukuttan/indic\_nlp\_library} \citep{kunchukuttan2020indicnlp} and English text processing uses the \textit{sacremoses}\footnote{https://github.com/alvations/sacremoses} package. 

We learn a BPE subword vocabulary using \textit{subword\_nmt} \citep{sennrich-etal-2016-neural} with 32K merge operations. We consider only those vocabulary items that have occurred at least 5 times in the training corpus. 

\subsubsection{Training and decoding} 

The models were trained using fairseq \cite{ott2019fairseq} on 8 V-100 GPUs. We optimized the cross entropy loss using the Adam optimizer with a label-smoothing of 0.1 and gradient clipping of 1.0. We use mixed precision training with Nvidia Apex\footnote{https://github.com/NVIDIA/apex}. We use an initial learning rate of 5e-4, 4000 warmup steps and the same learning rate annealing schedule as proposed in \citet{vaswani2017attention}. We use a global batch size of 262K tokens\footnote{We used a gradient update of 16 and per GPU batch size of 2048 tokens.}. We use early stopping with the patience set to 5 epochs. For decoding, we use a beam size of 5.

\subsection{Validation and Evaluation} We use the n-way parallel validation and testset provided by the above mentioned WAT2021 shared task. The dev and test sets contain 1000 and 2390 sentences per language respectively. Since there are multiple translation directions we sample 10\% of the data for each translation direction following \citet{aharoni-etal-2019-massively}. The n-way testset enables evaluation in all translation directions. We use BLEU as the evaluation metric computed using the \textit{sacrebleu} package \cite{post-2018-call}. For Indic-English\footnote{\scriptsize{BLEU+case.mixed+numrefs.1+smooth.exp+tok.13a+version.1.5.1}}, we use the the in-built, default \texttt{mteval-v13a} tokenizer. For En-Indic\footnote{\scriptsize{BLEU+case.mixed+numrefs.1+smooth.exp+tok.none+version.1.5.1}}, we first tokenize using the IndicNLP tokenizer before running sacreBleu.

\section{Research Questions}
\label{sec:research_q}

This work explores the following research questions

\begin{itemize}
\item How do multi-bridge models perform under different data sampling strategies?
\item Does fine-tuning the multi-bridge models for certain translation directions improve translation quality?
\item How does translation between non-English languages compare with English-centric translation. 
\end{itemize}

\subsection{Data Sampling Strategies}

We studied MB-M2M models trained under different data sampling conditions for non-English pairs. In all the cases mentioned below, all English-centric data is used during training: 

\begin{itemize}
\item \textbf{SamplePairs}: Use all data from some non-English centric language pairs. We randomly select the language pairs while ensuring that all the languages involved are spanned. In our experiments, we used 22 language pairs (out of the possible 90). The language pairs used (in both directions) are: bn-hi, bn-mr, bn-or, gu-pa, gu-te, hi-ta, kn-pa, kn-ta, ml-or, ml-te, mr-te. 
\item \textbf{SampleFraction}: Sample from all non-English  language pairs. We sample around 100k sentence pairs from each non-English pair so that the total non-English parallel corpora is roughly equal to the \textit{SamplePairs} method. 
\item \textbf{TrainAll}: All parallel data from all the non-English pairs are used.
\end{itemize}

\subsection{Baselines}

\begin{table*}[t]
\centering
\begin{tabular}{llllllllllll}
\toprule
Model           & bn   & gu   & hi   & kn   & ml   & mr   & or   & pa   & ta   & te   & AVG \\
\midrule
bilingual       & 22.8 & 32.2 & 39.2 & 24.3 & 26.6 & 26.9 & 25.7 & 36.2 & 28.0 & 25.9 & 28.8    \\
EC-M2O          & 27.7 & 38.4 & 41.8 & 34.2 & 32.6 & 31.8 & 32.4 & 41.0 & 31.2 & 33.5 & 34.5    \\
\midrule
\multicolumn{12}{l}{\textit{MB-M2M}}              \\
SamplePairs     & 28.8 & 39.3 & 42.3 & 34.8 & 33.7 & 32.3 & 32.8 & 41.4 & 32.0 & 34.9 & 35.2    \\
SampleFraction  & 28.6 & 38.7 & 41.9 & 34.9 & 33.7 & 32.6 & 33.1 & 41.3 & 32.0 & 35.0 & 35.2    \\
TrainAll        & 27.7 & 37.9 & 41.3 & 34.2 & 32.5 & 31.4 & 32.0 & 40.4 & 30.9 & 34.1 & 34.2    \\
\bottomrule
\end{tabular}
\caption{Results for translation into English from other languages (BLEU score).}
\label{tab:x2e_results}
\end{table*}

We also train many-to-one (M2O), one-to-many (O2M) and English-centric many-to-many (EC-M2M) (\textit{transformer\_vaswani\_wmt\_en\_de\_big} architecture). We use the M2O and O2M models for pivot baselines and EC-M2M for zeroshot baselines for evaluating translation between Indic languages. We also train bilingual baselines for all English-Indian, Indian-English directions using the small \textit{transformer\_iwslt\_de\_en} architecture and for 10 Indian language pairs using the base \textit{transformer} architecture. 

\begin{table*}[t]
\centering
\begin{tabular}{llllllllllll}
\toprule
Model          & bn   & gu   & hi   & kn   & ml   & mr   & or   & pa   & ta   & te   & AVG \\
\midrule
bilingual      & 13.5 & 22.4 & 36.2 & 13.5 & 11.7 & 16.1 & 14.3 & 30.5 & 12.9 & 12.6 & 18.4    \\
EC-O2M         & 14.2 & 23.4 & 34.9 & 16.8 & 13.2 & 17.6 & 17.2 & 30.4 & 12.3 & 15.0 & 19.5    \\
\midrule
\multicolumn{12}{l}{\textit{MB-M2M}}  \\
SamplePairs    & 15.0 & 25.7 & 37.2 & 18.3 & 14.1 & 18.8 & 18.3 & 32.9 & 13.5 & 16.3 & 21.0    \\
SampleFraction & 14.9 & 25.7 & 37.4 & 17.9 & 13.7 & 18.9 & 18.0 & 32.6 & 13.2 & 16.5 & 20.9    \\
TrainAll       & 14.1 & 24.7 & 36.8 & 18.2 & 13.6 & 18.1 & 17.9 & 32.2 & 12.9 & 15.9 & 20.4    \\
\bottomrule
\end{tabular}
\caption{Results for translation from English into other languages (BLEU score).}
\label{tab:e2x_results}
\end{table*}

\section{Results and Analysis}
\label{sec:results}

In this section, we discuss the results of our experiments. 

\subsection{Translation between non-English languages}

Table \ref{tab:x2y_results} shows the results of translation between non-English languages.

\subsubsection{Comparison with zeroshot and pivot  translation} The English-centric M2M translation model performs very poorly for zeroshot translation between Indian languages. The model always generates English output which is consistent with previous observations in literature. Zeroshot translation is particularly pathological for Indian languages since their scripts are totally different from the Latin English script - hence there is very little vocabulary sharing and the model exclusively generates English for translation between Indian languages. All the multi-bridge models perform better than the pivot model. We also find that the multilingual and pivot models are substantially better than bilingual models. To illustrate that, we show results for some non-English pairs in  Table \ref{tab:x2y_sample_pairs}.

\subsubsection{Data Sampling conditions} All three data sampling methods show roughly the same translation quality, suggesting that all the data is not needed for training  non-English directions. Particularly, \textit{SampleFraction} performs is slightly better than \textit{SamplePairs} and is equivalent to \textit{TrainAll}. 

\subsubsection{Indic-data only models} They are inferior to the models using English-centric data pointing to the need for the original English-centric data as well.

\subsubsection{Comparison with English-centric models} We compare translations from English to Indian languages with translations between Indian languages. The n-way nature of the WAT2021 testset make such a study possible. It is desirable that translation between Indian languages achieve the same quality as English-Indic languages since  multilingual model enable transfer learning and the Indic training corpus is extracted from the English-centric corpus. We compare the average translation quality metrics of: (1) English to Indian language directions, and (2) Indian-Indian language directions. We use multiple evaluations metrics to obtain a holistic comparison: BLEU (token-based) \cite{papineni-etal-2002-bleu}, chrF2 (character-based) \cite{popovic-2015-chrf} and LABSE cosine similarity (semantic similarity) \cite{feng2020languageagnostic}. Table \ref{tab:eval_comparison} shows that the English to Indian language translation shows a higher quality than Indic-Indic translation on all these metrics. 

A possible explanation for this gap is the nature of the testsets - most likely the English source sentence has been independently translated into all Indic languages. This might cause a semantic drift in the reference translations between the Indic languages and is a potential issue in n-way testsets. To check this hypothesis, we compute the semantic similarity between the source and target languages in the testset using cosine similarity between  LABSE representations of the sentences. We see that semantic similarity between Indic languages is roughly similar to that between English and Indian languages. Hence, We can discount any semantic drift in the testsets. 

\subsection{Translation to English}

Table \ref{tab:x2e_results} show the results. We observe that all the multilingual models are significantly better than the bilingual models. The multi-bridge models perform better than the English-centric many to one model. The SampledFraction and SamplePairs strategies outperform the TrainAll model. The TrainAll model sees more non-English language data training and hence its performance on English target might be degrading.  

\subsection{Translation from English}

Table \ref{tab:e2x_results} show the results. We observe that all the multilingual models are significantly better than the bilingual models. Further, it is important to note that the multi-bridge models significantly outperform the English-centric one to many model (about 1.5 BLEU points average across languages). Improving multilingual models when multiple target languages are involved has been a challenge and these results indicate that multibridge models can provide significant improvement in this scenario. The SampledFraction and SamplePairs strategies outperform the TrainAll strategy.

\begin{table*}[]
\centering
\begin{tabular}{lrrrrrrrrrrr}
\toprule
Model                            & bn   & gu   & hi   & kn   & ml   & mr   & or   & pa   & ta   & te   & AVG \\
\midrule
\multicolumn{12}{l}{\textit{Translation between non-English languages}}              \\
SamplePairs                      & 12.9 & 19.9 & 29.3 & 13.8 & 10.4 & 15.4 & 15.2 & 26.4 & 10.7 & 12.6 & 16.7    \\
SampleFraction                   & 12.3 & 19.9 & 29.9 & 13.7 & 10.4 & 15.1 & 15.0 & 26.5 & 10.2 & 12.6 & 16.6    \\
TrainAll                         & 12.3 & 19.7 & 29.3 & 13.6 & 10.3 & 15.1 & 15.0 & 25.8 & 10.2 & 12.5 & 16.4    \\

\midrule
\multicolumn{12}{l}{\textit{Translation into English from other languages}}              \\
SamplePairs     & 29.1 & 39.0 & 42.4 & 34.6 & 33.6 & 32.3 & 32.7 & 41.3 & 32.3 & 34.6 & 35.2    \\
SampleFraction  & 28.3 & 38.3 & 42.0 & 34.5 & 33.4 & 31.9 & 33.0 & 41.3 & 32.1 & 34.6 & 35.0    \\
TrainAll        & 28.1 & 37.5 & 41.4 & 34.0 & 32.3 & 32.0 & 32.1 & 40.3 & 31.5 & 33.7 & 34.3   \\

\midrule
\multicolumn{12}{l}{\textit{Translation from English into other languages}} \\
SamplePairs    & 15.3 & 25.6 & 37.2 & 18.1 & 13.7 & 18.7 & 18.9 & 32.9 & 13.4 & 16.4 & 21.0    \\
SampleFraction & 15.0 & 25.5 & 37.5 & 18.4 & 13.6 & 19.0 & 17.9 & 33.2 & 13.3 & 16.2 & 21.0    \\
TrainAll       & 15.3 & 25.8 & 37.7 & 18.8 & 13.9 & 19.6 & 18.8 & 33.3 & 13.9 & 16.5 & 21.4 \\
\bottomrule
\end{tabular}
\caption{Results for finetuning of multibridge models. For translation between Indic languages, each cell shows the average BLEU score for translating into corresponding language. The last column shows the average BLEU score for the model}
\label{tab:finetune_results}
\end{table*}

\subsection{Finetuning of multi-bridge models}

We finetune the multibridge models by continuing training on the finetuning dataset. We experiment with 3 finetuning scenarios: (1) non-English parallel data, (2) English-Any parallel data, (3) Any language to English parallel data.  We stop finetuning when we see no improvement on the validation set for 5 epochs. The results of finetuning of multibridge models are shown in Table \ref{tab:finetune_results}. Finetuning these models on non-English data alone does not yield any major gains.  Fine-tuning the model on X to English data only also does not result in any significant change. Finetuning the model on English to language data improves the TrainAll model by a BLEU point.

\section{Conclusion}
\label{sec:conclusion}

We conduct a systematic analysis of multibridge multilingual NMT models. Our results show that only a small amount of translation data from non-English pairs is sufficient to achieve best results with standard multilingual training. It is possible to train a single multilingual model to serve multiple translation directions. Further advances in multilingual learning are needed to achieve better transfer from English-centric directions to non-English directions. Particularly, when the non-English languages are related, better methods to utilize relatedness of these languages are required. 

\bibliography{anthology,custom}

\end{document}